\pdfoutput=1

\documentclass[11pt]{article}

\usepackage{naacl2021}

\usepackage{times}
\usepackage{latexsym}

\usepackage[T1]{fontenc}

\usepackage[utf8]{inputenc}

\usepackage{microtype}

\usepackage{graphicx}
\usepackage{booktabs}

\usepackage{fancyvrb}

\title{LAST at CMCL 2021 Shared Task: Predicting Gaze Data During Reading with a Gradient Boosting Decision Tree Approach}

\author{Yves Bestgen \\
  Laboratoire d'analyse statistique des textes - LAST \\
  Institut de recherche en sciences psychologiques \\
  Universit\'e catholique de Louvain \\
  Place Cardinal Mercier, 10 1348 Louvain-la-Neuve, Belgium \\
  \texttt{yves.bestgen@uclouvain.be} \\}

\begin{document}
\maketitle
\begin{abstract}
A LightGBM model fed with target word lexical characteristics and features obtained from word frequency lists, psychometric data and bigram association measures has been optimized for the 2021 CMCL Shared Task on Eye-Tracking Data Prediction. It obtained the best performance of all teams on two of the five eye-tracking measures to predict, allowing it to rank first on the official challenge criterion and to outperform all deep-learning based systems participating in the challenge.
\end{abstract}

\section{Introduction}

This paper describes the system proposed by the Laboratoire d’analyse statistique des textes (LAST) for the Cognitive Modeling and Computational Linguistics
(CMCL) Shared Task on Eye-Tracking Data Prediction. This task is receiving more and more attention due to its importance in modeling human language understanding and improving NLP technology \citep{hollenstein-etal-2019-cognival,Mishra2018}. 

As one of the objectives of the organizers is to “compare the capabilities of machine learning approaches to model and analyze human patterns of reading” (\url{https://cmclorg.github.io/shared\_task}), 
I have chosen to adopt a generic point of view with the main objective of determining what level of performance can achieve a system derived from the one I developed to predict the lexical complexity of words and polylexical expressions \citep{shardlow2021semeval}. That system was made up of a gradient boosting decision tree prediction model fed with features obtained from word frequency lists, psychometric data, lexical norms and bigram association measures. If there is no doubt that predicting lexical complexity is a different problem, one can think that the features useful for it also play a role in predicting eye movement during reading.  

The next section summarizes the main characteristics of the challenge. Then the developed system is described in detail. Finally, the results in the challenge are reported along with an analysis performed to get a better idea of the factors that affect the system performance.

\section{Data and Task}
The eye-tracking data for this shared task were extracted from the Zurich Cognitive Language Processing Corpus (ZuCo 1.0 and ZuCo 2.0, \citealp{Hollenstein18,HollensteinTZL20}). 
It contains gaze data for 991 sentences read by 18 participants during a normal reading session. The learning set consisted in 800 sentences and the test set in 191 sentences. 

The task was to predict five eye-tracking features, averaged across all participants and scaled in the range between 0 and 100, for each word of a series of sentences: (1) the total number of fixations (nFix), (2) the duration of the first fixation (FFD), (3) the sum of all fixation durations, including regressions (TRT), (4) the sum of the duration of all fixations prior to progressing to the right, including regressions to previous words (GPT), and (5) the proportion of participants that fixated the  word (fixProp). These dependent variables (DVs) are described in detail in \citet{Hollenstein21}. The submissions were evaluated using the mean absolute error (MAE) metric and the systems were ranked according to the average MAE across all five DVs, the lowest being the best.

As the DVs are of different natures (number, proportion and duration), their mean and variance are very different. The mean of fixProp is 21 times greater than that of FFD and its variance 335 times. Furthermore, while nFix and fixProp were scaled independently, FFD, GPT and TRT were scaled together. For that reason, the mean and dispersion of these  three measures are quite different: FFD must necessarily be less than or equal to TRT and GPT\footnote{The relation between TRT and GPT is not obvious to me since one can be larger or smaller than the other in a significant number of cases.}. These two factors strongly affect the importance of the different DVs in the final ranking.

\section{System}

\subsection{Procedure to Build the Models}

The regression models were built by the 2.2.1 version of the LightGBM software \citep{LightGBM}, a well-known implementation of the gradient boosting decision tree approach. This type of model has the advantage of not requiring feature preprocessing, such as a logarithmic transformation, since it is insensitive to monotonic transformations, and of including many parameters allowing a very efficient overfit control. It also has the advantage of being able to directly optimize the MAE.

Sentence preprocessing and feature extraction as well as the post-processing of the LightGBM predictions were performed using custom SAS programs running in SAS University (still freely available for research at \url{https://www.sas.com/en_us/software/university-edition.html)}. Sentences were first lemmatized by the TreeTagger \citep{schmid1994treetagger} to get the lemma and POS-tag of each word. Special care was necessary to match the TreeTagger tokenization with the Zuco original one. Punctuation marks and other similar symbols (e.g., "(" or "\$") were simply disregarded as they were always bound to a word in the tokens to predict. The attribution to the words of the values on the different lists was carried out in two stages: on the basis of the spelling form when it is found in the list or of the lemma if this is not the case. 

The features used in the final models as well as the LightGBM parameters were optimized by a 5-fold cross validation procedure, using the sentence and not the token as the sampling unit. The number of boosting iterations was set by using the LightGBM early stopping procedure which stops training when the MAE on the validation fold does not improve in the last 200 rounds. The predicted values which were outside the [0, 100] interval were brought back in this one, which makes it possible to improve the MAE very slightly.

\subsection{Features}
To predict the five DVs, five different models were trained. The only differences between them were in the LightGBM parameters. There were thus all based on exactly the same features, which are described below.
\paragraph{Target Word Length.}
The length in characters of the preceding word, the target word and the following one.
\paragraph{Target Word Position.}
The position of the word in the sentence encoded in two ways: the rank of the word going from 1 to the sentence total number of words and the ratio between the rank of the word and the total number of words.
\paragraph{Target Word POS-tag and Lemma.}
The POS-tag and lemma for the target word and the preceding one.

\paragraph{Corpus Frequency Features.}
Frequencies in corpora of words were either calculated from a corpus or extracted from lists provided by other researchers. The following seven features have been used:
\begin{itemize}
\item The (unlemmatized) word frequencies in the British National Corpus (BNC, \url{http://www.natcorp.ox.ac.uk/}).
\item The Facebook frequency norms for American English and British English in \citet{HER17}.
\item The Rovereto Twitter Corpus frequency norms \citep{HER17}.
\item The USENET Orthographic Frequencies from \citet{SHA06}.
\item The Hyperspace Analogue to Language (HAL) frequency norms provided by \citep{BAL07} for more that 40,000 words.
\item The frequency word list derived from Google's ngram corpora available at \url{https://github.com/hackerb9/gwordlist}.
\end{itemize}

\paragraph{Features from Lexical Norms.}
The lexical norms of Age of Acquisition and Familiarity were taken from the Glasgow Norms which contain judges' assessment of 5,553 English words \citep{SCO19}.

\paragraph{Lexical Characteristics and Behavioral Measures from ELP.}
Twenty-three indices were extracted from the English Lexicon Project (ELP, \citealp{BAL07,YAR08}), a database that contains, for more than 40,000 words, reaction time and accuracy during lexical decision and naming tasks, made by many participants, as well as lexical characteristics (\url{https://elexicon.wustl.edu/}).
Eight indices come from the behavioral measures, four for each task: average response latencies (raw and standardized), standard deviations, and accuracies.
Fourteen indices come from the “Orthographic, Phonological, Phonographic, and Levenshtein Neighborhood Metrics” section of the dataset. These are all the metrics provided except Freq\_Greater, Freq\_G\_Mean, Freq\_Less, Freq\_L\_Mean, and Freq\_Rel. These are variables whose initial analyzes showed that they were redundant with those selected. The last feature is the average bigram count of a word.

\paragraph{Bigram Association Measures.}
These features indicate the degree of association between the target word and the one that precedes it according to a series of indices calculated on the basis of the frequency in a reference corpus (i.e., the BNC) of the bigram and that of the two words that compose it, using the following association measures (AMs): pointwise mutual information and t-score \citep{church-hanks-1990-word}, z-score \citep{BER73}, log-likelihood Chi-square test \citep{dunning-1993-accurate}, simple-ll \citep{EVE09}, Dice coefficient \citep{KIL14} and the two delta-p \citep{KYL18}. Most of the formulas to compute these AMs are also provided in \citet{EVE09} and in \citet{PEC10}. As these features mix together the assets of both collocations (by using association scores) and ngrams (by using contiguous pairs of words), \citet{BG14} refer to them as \textit{collgrams}. They make it possible not to rely exclusively on the frequency of the bigram in the corpus, which can be misleading because a bigram may be observed frequently, not because of its phraseological nature, but because it is made up of very frequent words \citep{Bes2018}. Conversely, a relatively rare bigram, composed of rare words, may be typical of the language. Since word frequency is already accounted for by the corpus frequency features, it was desirable to employ indices that reduce the impact of this factor. Originating in works in lexicography and foreign language learning \citep{church-hanks-1990-word,DUR09,BES17, BE19rfla}, they have recently shown their usefulness in predicting the lexical complexity of multi-word expressions \citep{BES21}. In the present case, it is assumed that these indices can serve as a proxy of the next word predictability \citep{KLI04}.

\paragraph{Feature coverage.}
Some words to predict are not present in these lists and the corresponding score is thus missing. Based on the complete dataset provided by the organizers, it happens in:
\begin{itemize}
\item 1\% (Google ngram) to 17\% (Facebook and Twitter) of the tokens for the corpus frequency features,
\item 9\% for the ELP Lexical Characteristics, but a few features have as much as 41\% missing values,
\item 11\% for the ELP Behavioral Measures,
\item 18\% for the Bigram AMs.
\end{itemize}

In total, sixteen tokens have missing values for all these features (Corpus Frequency, Lexical Characteristics and Behavioral Measures from ELP, and Bigram Association Measures). These tokens have however received values for the length and position features. All the missing values were handled by LightGBM default procedure.

\begin{table}
\begin{center}
\begin{tabular}{lrr}
\toprule
Parameters     &  \multicolumn{1}{c}{Run 1}      &   \multicolumn{1}{c}{Run 2} \\ \midrule
bagging\_fraction & 0.66 & 0.70 \\ 
bagging\_freq & 5 & 5 \\ 
feature\_fraction & 0.09 & 0.85 \\ 
learning\_rate & 0.0095 & 0.0050 \\ 
max\_depth & 11 & no limit \\ 
max\_bin & 64 &  64 \\ 
min\_data\_in\_bin & 2 & 5 \\ 
max\_leaves & 11 & 30 \\ 
min\_data\_in\_leaf & 7 & 5 \\ 
n\_iter & 4800 & (see text) \\ \bottomrule
\end{tabular}
\caption{LightGBM parameters for the first two runs.}
\end{center}
\end{table}

\begin{table*}
\centering
\begin{tabular}{llrrrrrr}
\toprule
Team & Run & \multicolumn{1}{c}{Mean} & \multicolumn{1}{c}{nFix} & \multicolumn{1}{c}{FFD} & \multicolumn{1}{c}{GPT} & \multicolumn{1}{c}{TRT} & \multicolumn{1}{c}{fixProp} \\ \midrule
LAST & 3 & \bf 3.8134 & 3.879 & \bf 0.655 & 2.197 & 1.524 & \bf 10.812 \\
LAST & 2 & 3.8159 & 3.886 & 0.655 & 2.199 & 1.523 & 10.817 \\
TALEP & 1 & 3.8328 & \bf 3.761 & 0.662 & \bf 2.180 & \bf 1.486 & 11.076 \\
LAST & 1 & 3.8664 & 3.943 & 0.662 & 2.237 & 1.545 & 10.944 \\
TorontoCL & 2 & 3.9287 & 3.944 & 0.671 & 2.227 & 1.516 & 11.286 \\ \bottomrule
\end{tabular}
\caption{Performance (MAE) for the five best runs submitted to the challenge. Best scores are bolded.}
\end{table*}

\section{Analyses and Results}
\subsection{Models Submitted to the Challenge} 
During the test phase, teams were allowed to submit three runs. My three submissions were all based on the features described above, the only differences between them resulting from changes in the LigthGBM parameters. They were set at their default values except those shown in Table 1. The official performances of the top five challenge submissions are given in Table 2.

The first submission was based on the parameters selected during the development phase. They were identical for the five DVs. For the other two submissions, a random grid search coded in python was used to try optimizing the parameters independently for each DV. The parameter space for this first random search is provided in Appendix A. As the measure of the challenge is the MAE averaged across the five DVs and as the system MAE for fixProp was up to 15 times higher than that of the other DVs, the optimized parameters for this variable were selected. Additional analyzes showed that they also made it possible to improve performance on the four other DVs. Their values are given in Table 1. Certain initial choices were only slightly modified. The value of other parameters such as the maximum number of leaves and the feature fraction were markedly increased, suggesting that the risk of overfit was relatively low (see \url{https://lightgbm.readthedocs.io/en/latest/Parameters-Tuning.html}). In this system, the number of iterations was optimized (thanks to the early stopping procedure) for each DV and sets at the fourth highest value: 3,740 for nFix, 3,829 for TRT, 2,861 for GPT, 3,497 for FFD, and 3,305 for fixProp.

\begin{table*}
\centering
\begin{tabular}{lrrrrrrr}
\toprule
Models & \multicolumn{1}{c}{MAE} & \multicolumn{1}{c}{\%MAE} & \multicolumn{1}{c}{\%nFix} & \multicolumn{1}{c}{\%FFD} & \multicolumn{1}{c}{\%GPT} & \multicolumn{1}{c}{\%TRT} & \multicolumn{1}{c}{\%fixProp} \\ \midrule
W/o behavioral data & 3.849  & -0.93  & \bf -0.69  & \bf -1.30  & -0.75  & -0.78  & -1.05 \\
W/o ELP charact.  & 3.859  & -1.19  & \bf -0.54  & -1.36  & -0.95  & -0.59  & \bf -1.55 \\
W/o frequencies  & 3.880  & -1.74  & \bf -1.38  & -1.68  & \bf -1.88  & -1.55  & -1.87 \\
W/o bigram AM  & 3.881  & -1.78  & -2.05  & \bf -2.32  & \bf -1.39  & -1.94  & -1.70 \\
W/o length feat.  & 3.979  & -4.35  & \bf -5.95  & \bf -2.92  & -3.17  & -4.43  & -4.08 \\
W/o position feat.  & 4.095  & -7.39  & -7.68  & -4.44  & \bf-22.88  & -7.48  & \bf -4.30 \\ \midrule
RMSE optimization  & 3.847  & -0.87  & -0.43  & \bf 0.46  & \bf -4.73  & -0.09  & -0.43 \\
Default Param + MAE  & 3.902  & -2.32  & -2.34  & \bf -1.54  & \bf -3.52  & -2.12  & -2.15 \\
Default Param + RMSE  & 4.141  & -8.59  & -7.67  & -7.65  & \bf -12.62  & \bf -7.43  & -8.31 \\ \midrule
Linear Regression  & 4.268  & -10.64 & -9.04 & \bf -7.88 & \bf -24.09 & -9.47 & -8.26 \\
LGBM on Length + Position & 4.219  & -10.63  & -10.7  & -11.4  & \bf -8.18  & \bf -12.1  & -10.85 \\ \bottomrule
\end{tabular}
\caption{Performance (MAE) of different system versions and deviation (\%) from the best run ($MAE = 3.813$). Minimum and maximum values across DVs for each row are bolded.}
\end{table*}

For the third submission, a new round of random optimization was conducted by evaluating parameter values close to those selected for Run 2, independently for each DV. As it only got slightly better performance than Run 2, these parameter values are not shown to save space.

As shown in Table 2, Runs 2 and 3 ranked at the first 2 places of the challenge. This result was largely due to their better performance for fixProp since the TALEP system, second in the challenge, achieved significantly better performance for three of the five DVs, but these have less impact on the official measurement. An analysis, carried out after the end of the challenge, showed that the system would not have been more effective (average MAE of 3.8138) if, during the first optimization step, a specific model for each DV had been selected.

Using Pearson's linear correlation coefficient as a measure of effectiveness, which is unaffected by the differences in means and variability between the five DVs, Run 3 obtains an average r of 0.812 on the test set (min = 0.792 for GPT; max = 0.838 for fixProp). This value is relatively high, but it can only really be interpreted by taking into account the reliability of the average real eye-tracking feature values.

\subsection{Feature Usefulness}
The first part of Table 3 presents the main results of an ablation procedure aimed at examining the impact of the different types of features on the system performance. It gives the average MAE as well as the difference in percentage between each system and the best run for the average MAE and for the five DVs. It must be first stressed that all features based on lemmas and POS-tag, the two Glasgow norms and the length of the token that follows the target are useless for predicting the test set since without them the system achieves a MAE of 3.8134. They are thus discarded in all the ablation analyses. The target's positions in the sentence and the length features are clearly essential. Among the features resulting from corpora and behavioral data, it is the bigram association measures and the frequencies in the corpora that are the most useful.

Generally speaking, the feature sets have comparable utility for all DVs. However, we observe that the position in the sentences is particularly important for predicting GPT while the length of the target is more useful for nFix.

The second part of Table 3 presents an analysis of the utility of optimizing the LightGBM parameters, based on the best system. Optimizing RMSE instead of MAE is especially penalizing for GPT. Using the default values of the LightGBM parameters is particularly penalizing when RMSE is the criterion.

A final question concerns the benefits of employing LightGBM instead of another regression algorithm when the proposed features are used. To try to provide at least a partial answer, I trained a multiple linear regression model on the basis of the features used, while adding for each feature, for which the calculation was possible, a second feature containing the logarithm of the initial value. I replaced the missing data with 0, which is probably not optimal. A stepwise regression procedure with a threshold to enter sets at $p = 0.01$ and a threshold to exit sets at $p = 0.05$ was employed to construct for each DV a model on the learning set and apply it to the test set. The results obtained are given in the second to last row of Table 3. The performances are clearly less good. It is even worse than the performance level of a LightGBM model based only on the length and position features (see the last row of Table 3). This regression system would have been ranked 10th in the challenge.

\section{Conclusion}
The system proposed for the 2021 CMCL Shared Task on Eye-Tracking Data Prediction was particularly effective, obtaining the first place in the challenge, but it must be kept in mind that the system that came second is superior to it for three of the five DVs. The analyzes carried out to understand its pros and cons indicate that optimizing the LightGBM parameters is quite beneficial to it as well as the different sets of features derived from corpora and behavioral data, including bigram AMs which, to my knowledge, have never been employed for this type of task.

It would have been interesting to relate these observations to the psycholinguistic literature on the factors that influence eye fixations, but this is unfortunately not possible here, for lack of space. In addition, this would first require deepening the ablation analyzes by simultaneously considering several feature sets. For instance, the lack of usefulness of the POS-tags could simply result from the links (at least partial) between them and the frequency and length of the tokens. Likewise, some of the bigram AMs are relatively sensitive to the frequency of the words that compose them (e.g., the t-score favors frequent bigrams which are usually composed of frequent words). It is thus highly probable that some of the features in the different sets (frequencies, behavioral data...) are redundant and can be removed without impairing the performance of the system. This is a potential development path.

\section*{Acknowledgements}
The author wishes to thank the organizers of this shared task for putting together this valuable event and the reviewers for their very constructive comments. He is a Research Associate of the Fonds de la Recherche Scientifique - FNRS (F\'ed\'eration Wallonie Bruxelles de Belgique). Computational resources were provided by the supercomputing facilities of the UCLouvain (CISM/UCL) and the Consortium des Equipements de Calcul Intensif en F\'ed\'eration Wallonie Bruxelles (CECI).

\bibliography{SemEval-2021Task1}
\bibliographystyle{acl_natbib}

\appendix
\section{Appendix}

At the request of a reviewer, the parameter space for the first random search is provided below. Those for the second random search are not provided as they did not allow to really improve the performances.
\begin{Verbatim}[fontsize=\small]
param_grid = {
 'max_bin': [16,32,48,64,80,96,112,128,
    160,192,224,256],
 'min_data_in_bin':[2,3,4,5,6,8,10,12,
    15,20],
 'num_leaves': [4,5,6,7,8,9,10,11,12,13,
    15,18,21,25,30],
 'learning_rate': [0.005,0.007,0.009,
    0.011,0.014,0.018,0.022,0.026,0.03,
    0.035,0.05],
 'min_data_in_leaf': [2,3,4,5,6,7,8,9,10,
    11,12,13,15,18,21,25,30],
 'max_depth': [3,4,5,6,7,8,9,10,11,12,13,
    -1],
 'feature_fraction': list(np.linspace(
    0.01, 0.90, 91)),
 'bagging_freq': list(range(3, 7, 1)),
 'bagging_fraction': list(np.linspace(
    0.50, 0.90, 9))
}
\end{Verbatim}

\end{document}